\documentclass[11pt]{article}

\usepackage[preprint]{acl}
\usepackage{times}
\usepackage{latexsym}
\usepackage[T1]{fontenc}
\usepackage[utf8]{inputenc}
\usepackage{microtype}
\usepackage{inconsolata}
\usepackage{graphicx}
\usepackage{booktabs}
\usepackage{amsmath}
\usepackage{array}
\usepackage{tabularx}
\usepackage{xspace}
\usepackage{url}

\graphicspath{{figures/}}

\newcommand{\method}{Evidence Sufficiency Boundary Training\xspace}

\newcommand{\czer}{\textsc{C0}\xspace}
\newcommand{\cone}{\textsc{C1}\xspace}
\newcommand{\ctwo}{\textsc{C2}\xspace}
\newcommand{\cthree}{\textsc{C3}\xspace}

\title{Learning Evidence Sufficiency Boundaries for Selective Answering in Grounded Multi-Hop QA}

\author{Haruto Sato \and Yuki Tanaka \and Ren Nakamura\\
Aoi Kobayashi \and Mei Ito\\
\textnormal{Independent Researcher}}

\begin{document}
\maketitle

\begin{abstract}
Grounded question answering systems should answer only when the supplied evidence supports the answer. In multi-hop QA, this requirement is difficult because partial evidence can make an unsupported answer appear plausible. We study selective answering through evidence sufficiency boundaries: for the same question, a model should abstain under unsupported or partially supported context, answer when the context first becomes sufficient, and keep the answer stable when redundant evidence is added. We introduce \method, a generation-native training framework that constructs ordered evidence chains and supervises the abstain-to-answer transition directly. The method combines level supervision, a boundary flip margin, post-boundary stability, and answer recall protection. We build evidence chains from HotpotQA, 2WikiMultiHopQA, and MuSiQue, then evaluate models with chain metrics, raw QA utility, and unsupported-answer rates on external non-answerable sets. With Qwen2.5-3B-Instruct and LoRA adaptation, \method gives the strongest boundary localization among the tested systems, with flip accuracy of 0.807 compared with 0.781 for a token-level abstention baseline. It also achieves the lowest overall unsupported-answer rate on external non-answerable evaluation, 0.095 compared with 0.101 for the same baseline, while retaining competitive raw QA F1. The results show that grounded selective answering improves when training marks the evidence level where refusal should give way to answering.
\end{abstract}

\section{Introduction}

Large language models can answer many factual questions after instruction tuning and scaling, but they still produce unsupported statements when the input context is incomplete or misleading \citep{brown2020language,ouyang2022training,touvron2023llama,openai2023gpt4,qwen2025qwen25}. Retrieval-augmented and context-grounded systems reduce this risk by supplying external passages at inference time \citep{lewis2020rag,karpukhin2020dense,guu2020realm,izacard2021leveraging,borgeaud2022retro}. The remaining failure is operationally important: a system can receive context, follow the requested answer format, and still answer with content that the context does not support \citep{maynez2020faithfulness,ji2023survey,dziri2022faithdial,lin2022truthfulqa,huang2025survey}.

Grounded multi-hop QA exposes this problem sharply. Benchmarks such as HotpotQA, 2WikiMultiHopQA, and MuSiQue require evidence from multiple facts or documents \citep{yang2018hotpotqa,ho2020constructing,trivedi2022musique}. A single supporting fact can mention the right entity, relation, or topic while leaving the answer underdetermined. Standard answer metrics such as exact match and token F1 measure whether the final string matches the gold answer, but they do not measure whether the system should have answered under the available evidence. This gap has motivated context-faithfulness datasets and hallucination corpora for retrieval-augmented generation \citep{joren2025sufficient,niu2024ragtruth,ming2024faitheval,li2023halueval}.

The central object in this paper is the evidence sufficiency boundary. For one question family, context can be arranged from unsupported evidence, to partial evidence, to minimally sufficient evidence, and then to redundant evidence. The desired model behavior is simple: abstain before the boundary, answer at the first sufficient context, and keep answering after the boundary if the additional evidence preserves support. This view gives selective answering a location inside the evidence chain. It also gives a more precise test than measuring refusal frequency alone.

\begin{figure*}[t]
    \centering
    \includegraphics[width=\textwidth]{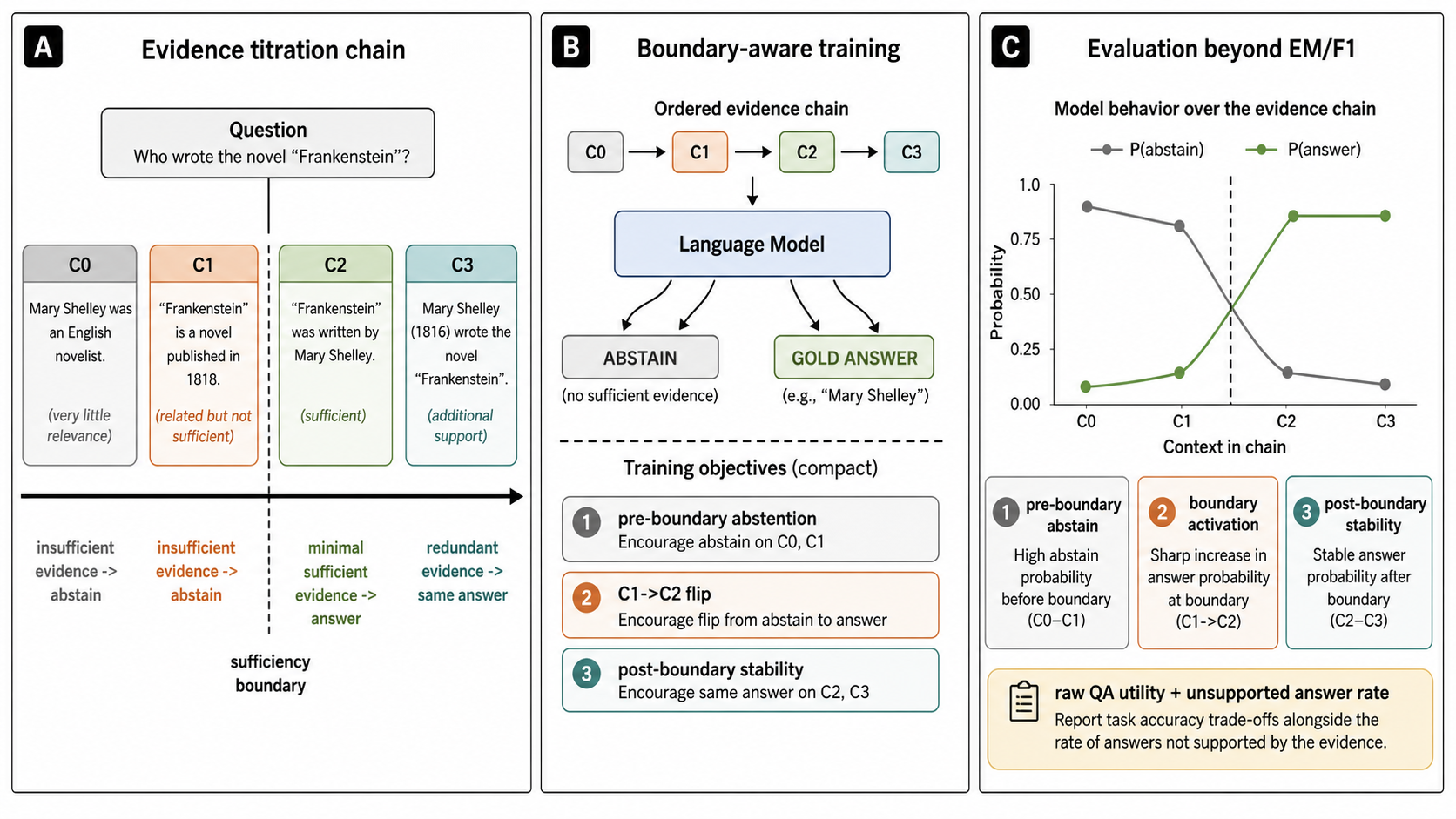}
    \caption{Overview of \method. Each question is converted into an ordered evidence chain. The model learns to abstain before the sufficiency boundary, activate the answer at the first sufficient context, and keep the answer stable after redundant evidence is added. Evaluation therefore measures boundary behavior in addition to raw QA utility and unsupported answering.}
    \label{fig:main}
\end{figure*}

We propose \method, a training framework for grounded multi-hop QA. The framework constructs four-level evidence chains for each question family. The first two levels are insufficient; the third level is the minimal sufficient context; the fourth level adds redundant support or benign extra context. A single generation model is trained to output \texttt{<ABSTAIN>} on insufficient levels and the gold short answer on sufficient levels. The training objective adds margin terms that make the transition from \cone to \ctwo explicit and penalize collapse after \ctwo. Figure~\ref{fig:main} summarizes the data, training objective, and evaluation.

This formulation connects three lines of prior work. First, unanswerable QA and refusal tuning teach models to avoid answering when the question or the model's parametric knowledge is unreliable \citep{rajpurkar2018squad2,zhang2024rtuning,yin2023llmsknow,zhao2024knowing}. Second, selective prediction and calibration provide tools for abstention under uncertainty \citep{chow1970optimum,elyaniv2010foundations,geifman2017selective,guo2017calibration,hendrycks2017baseline,kamath2020selective}. Third, token-level abstention methods such as SEAL introduce explicit reject probability into generation and decoding \citep{huang2025seal}. Our setting keeps the grounded evidence context fixed as the object of study and asks where the abstain-to-answer transition occurs.

Experiments use the same source question families for all methods. The main training split contains 2,400 families drawn evenly from HotpotQA, 2WikiMultiHopQA, and MuSiQue. We compare \method with answer-only training, a token-level abstention baseline adapted from SEAL, and a question-level refusal baseline adapted from R-Tuning. \method reaches the highest flip accuracy on the evidence chain, 0.807, while maintaining a pre-boundary abstention rate of 0.934. On external non-answerable evaluation, it obtains the lowest overall unsupported-answer rate, 0.095. SEAL remains a strong baseline, with slightly higher raw QA F1 and stronger post-boundary stability. The current evidence supports a bounded claim: evidence-boundary supervision improves boundary localization and external unsupported-answer suppression while preserving competitive answer utility.

The paper makes four contributions:
\begin{itemize}
    \item We formulate grounded selective answering as evidence sufficiency boundary learning.
    \item We introduce a titration-style data construction pipeline that converts grounded multi-hop QA examples into ordered evidence chains.
    \item We propose a generation-native training objective with pre-boundary abstention, boundary activation, post-boundary stability, and recall protection.
    \item We report boundary-sensitive evaluation alongside raw QA and unsupported-answer metrics, showing behavior that standard EM/F1 hides.
\end{itemize}

\section{Evidence Sufficiency Boundaries}

\subsection{Grounded Selective Answering}

Let $q$ be a question, $a$ a short gold answer, and $c$ a context supplied to the model. A grounded QA system should produce $a$ only when $c$ contains enough evidence to support $a$. Otherwise, the system should output a designated abstention response. We write the model output as
\begin{equation}
    y \in \{a, r\},
\end{equation}
where $r$ denotes the abstention token used during training and decoding.

The usual supervised QA setting provides a positive pair $(q,c,a)$. It rarely specifies whether smaller, partial, or perturbed contexts would still support $a$. This missing structure matters for multi-hop questions. If an answer depends on two facts, a context containing one fact should usually produce abstention. A context containing both facts should produce the answer. A context with both facts plus a distractor should preserve the answer unless the extra material invalidates the support.

\subsection{Boundary Definition}

For each question family, we construct an ordered chain
\begin{equation}
    \mathcal{C}(q)=\{c_0,c_1,c_2,c_3\}.
\end{equation}
The chain is ordered by evidence sufficiency. Contexts $c_0$ and $c_1$ are insufficient. Context $c_2$ is the first context judged sufficient for the gold answer. Context $c_3$ contains sufficient evidence with additional support or benign redundancy. The sufficiency boundary is
\begin{equation}
    k^\star=\min\{k: c_k \text{ supports } a\}.
\end{equation}
In our four-level construction, $k^\star=2$ by design after validation.

This definition focuses the model on a local transition. The model is correct on a family only if it keeps the pre-boundary contexts below the answer threshold and moves the sufficient contexts above it. A model that answers every level has high answer recall and poor boundary behavior. A model that abstains on every level has low hallucination and poor utility. A boundary-aware model must separate those two failures within the same question family.

\subsection{Metrics}

We use four chain metrics. Pre-boundary abstention is the fraction of \czer and \cone cases where the model abstains. Boundary activation is the fraction of \ctwo cases where the answer score is positive. Post-boundary stability is the fraction of \cthree cases where the answer score remains positive without a large drop from \ctwo. Flip accuracy is the fraction of families where the first positive answer score occurs at $k^\star$.

These metrics are reported with raw QA EM/F1 and unsupported-answer rates. Raw QA measures answer utility on standard answerable examples \citep{rajpurkar2016squad,joshi2017triviaqa,kwiatkowski2019natural}. Unsupported-answer rate measures whether generated answers are supported by the provided context. We report it both for answerable raw QA examples and for filtered external non-answerable examples.

\section{Evidence Chain Construction}

\subsection{Source Benchmarks}

The chain construction starts from grounded multi-hop QA benchmarks. HotpotQA contains questions with sentence-level supporting facts over Wikipedia passages \citep{yang2018hotpotqa}. 2WikiMultiHopQA adds multi-hop templates and explicit reasoning categories over Wikidata and Wikipedia \citep{ho2020constructing}. MuSiQue composes multi-hop questions from single-hop questions while reducing shortcut artifacts \citep{trivedi2022musique}. These datasets provide a useful base because the answer often depends on a small set of support facts embedded in a larger context.

The formal training split contains 2,400 question families: 800 from each benchmark. Each family has four contexts, yielding 9,600 chain rows. The answer-positive branch uses the same family source and contains 4,800 positive examples from \ctwo and \cthree. All baselines in the main comparison are exported from the same family set. This design keeps the comparison focused on supervision format and training objective.

\subsection{Four Context Levels}

Each family is converted into four context levels:
\begin{itemize}
    \item \czer: unsupported or weakly related context. The context should not identify the gold answer.
    \item \cone: partial support. The context contains topic or entity overlap, but it lacks a required bridge fact or disambiguating fact.
    \item \ctwo: minimal sufficient context. The context contains the evidence needed to support the gold answer.
    \item \cthree: sufficient context with redundancy. The context preserves the \ctwo support while adding extra support or benign distractors.
\end{itemize}

The construction uses supporting-fact metadata when available and applies programmatic filtering before validation. Candidate chains are rejected when a pre-boundary context leaks the answer, when the minimal sufficient context fails to support the answer, or when the redundant context changes the answer. This validation step is essential because a chain with a wrong boundary teaches the model an inconsistent policy.

\subsection{Baseline Exports}

The same chain families are exported into different training views. Answer-only training receives only \ctwo and \cthree positives. The token-level abstention baseline receives all four levels with \czer and \cone mapped to \texttt{[REJ]} and \ctwo and \cthree mapped to the gold answer. The question-level refusal baseline receives all four levels as sure/unsure supervision and an answer branch with the same positive budget. Table~\ref{tab:data_views} summarizes the training views.

\begin{table*}[t]
\centering
\small
\begin{tabularx}{\textwidth}{l l l X}
\toprule
Training view & Chain rows & Positive answer rows & Target behavior \\
\midrule
\method & 9,600 & 4,800 & Abstain on \czer/\cone, answer on \ctwo/\cthree, and enforce the \cone-to-\ctwo transition. \\
Answer-only training & 0 & 4,800 & Learn ordinary answer generation from sufficient contexts only. \\
Token-level abstention baseline & 9,600 & Included in chain rows & Map insufficient levels to \texttt{[REJ]} and sufficient levels to the gold answer with abstention-aware decoding. \\
Question-level refusal baseline & 9,600 & 4,800 & Learn sample-level sure/unsure decisions and use a calibrated gate before answer generation. \\
\bottomrule
\end{tabularx}
\caption{Training views derived from the same evidence-chain families. The source families are matched across methods; the difference is the supervision format.}
\label{tab:data_views}
\end{table*}

\section{\method}

\subsection{Model Interface}

The model receives an instruction, a question, and a context. It generates a short answer or \texttt{<ABSTAIN>}. We use a single generation path for both decisions. This keeps the method close to ordinary supervised fine-tuning and avoids a separate classifier at inference time.

For each context $c_i$, let $x_i$ be the formatted prompt containing $q$ and $c_i$. Let $r$ be the abstention token and $a_{1:p}$ the first $p$ answer tokens. We define an answer-vs-abstain score
\begin{equation}
    s_i = \log p_\theta(a_{1:p}\mid x_i) - \log p_\theta(r\mid x_i).
\end{equation}
In the simplest setting, $p=1$ and the score compares the first answer token against the abstention token at the generation boundary. Prefix scoring extends this comparison to a short answer prefix.

\subsection{Level Supervision}

The level objective trains the output string for each context:
\begin{equation}
\mathcal{L}_{\mathrm{level}} =
    \sum_{i < k^\star} \mathrm{CE}(r \mid x_i)
    + \sum_{i \ge k^\star} \mathrm{CE}(a \mid x_i).
\end{equation}
This objective gives the model direct generation targets. It does not by itself guarantee a clean boundary, since independent CE losses can still produce weak margins or unstable scores across adjacent levels.

\subsection{Boundary Losses}

The pre-boundary loss keeps insufficient contexts below the answer threshold:
\begin{equation}
    \mathcal{L}_{\mathrm{pre}} =
    \sum_{i<k^\star} \max(0, s_i + m_{\mathrm{pre}}).
\end{equation}
The flip loss makes the transition from \cone to \ctwo explicit:
\begin{equation}
    \mathcal{L}_{\mathrm{flip}} =
    \max(0, s_1 + m_{\mathrm{flip}})
    + \max(0, m_{\mathrm{flip}} - s_2).
\end{equation}
The post-boundary stability loss prevents the answer score from collapsing after redundant evidence is added:
\begin{equation}
    \mathcal{L}_{\mathrm{stab}} =
    \max(0, (s_2 - s_3) - \delta).
\end{equation}
The recall protection term keeps answerable QA examples from drifting toward abstention:
\begin{equation}
    \mathcal{L}_{\mathrm{recall}} =
    \max(0, m_{\mathrm{recall}} - s_{\mathrm{qa}}).
\end{equation}

The full training objective is
\begin{align}
\mathcal{L} =
&\lambda_{\mathrm{level}}\mathcal{L}_{\mathrm{level}}
+\lambda_{\mathrm{pre}}\mathcal{L}_{\mathrm{pre}}
+\lambda_{\mathrm{flip}}\mathcal{L}_{\mathrm{flip}} \nonumber\\
&+\lambda_{\mathrm{stab}}\mathcal{L}_{\mathrm{stab}}
+\lambda_{\mathrm{recall}}\mathcal{L}_{\mathrm{recall}}.
\end{align}
The objective is local to the evidence chain. It asks the model to place the answer threshold at the minimal sufficient context and to keep the same answer available after that point.

\subsection{Relation to Baselines}

Answer-only fine-tuning learns the answer distribution on sufficient contexts and receives no signal for pre-boundary abstention. Question-level refusal tuning adds an unsure state at the sample level, which helps when the main problem is unknown or unanswerable questions \citep{zhang2024rtuning}. Token-level abstention learning assigns probability mass to a rejection token when individual target tokens are unreliable and can use that probability in decoding \citep{huang2025seal}. \method uses the evidence chain as the supervision unit. Its score compares answer activation with abstention at adjacent evidence levels.

\section{Experimental Setup}

\subsection{Models and Training}

All experiments use Qwen2.5-3B-Instruct as the backbone \citep{qwen2025qwen25}. We adapt the model with LoRA \citep{hu2022lora} using the Hugging Face Transformers and PEFT implementations \citep{wolf2020transformers,mangrulkar2022peft}. The LoRA rank is 16, the scaling factor is 32, and dropout is 0.05. Training runs for 400 update steps with bf16 mixed precision, a maximum context budget of 1,400 characters, and a short-answer instruction. Greedy decoding is used for the main method and the answer-only baseline. The token-level abstention baseline uses its abstention-aware beam decoding with beam size 4, matching the corrected comparison protocol.

\subsection{Baselines}

\paragraph{Answer-only training.}
This baseline receives only sufficient contexts from \ctwo and \cthree. It tests whether ordinary QA supervision can learn selective behavior without negative evidence levels.

\paragraph{Token-level abstention learning.}
This baseline follows SEAL's core design: insufficient rows map to a rejection token, sufficient rows map to the gold answer, and decoding uses the learned rejection probability to penalize uncertain continuations \citep{huang2025seal}. The adapted baseline uses the same source families as \method.

\paragraph{Question-level refusal tuning.}
This baseline follows the R-Tuning direction of training a model to express uncertainty and refuse unknown questions \citep{zhang2024rtuning}. We match the positive answer budget by pairing the refusal branch with \ctwo/\cthree QA examples from the same family source.

\subsection{Evaluation}

We report three groups of metrics. Chain metrics measure whether the model has learned the evidence boundary. Raw QA metrics measure ordinary answer utility on held-out answerable examples. Unsupported-answer rates measure grounding reliability. External evaluation uses filtered non-answerable examples derived from the same benchmark families but held out from the training chains. A generated answer is counted as unsupported when it is not supported by the supplied context under the judge protocol used in the project evaluation.

\section{Main Results}

For compact tables, ESBT denotes \method. SEAL-style denotes the token-level abstention baseline, and R-Tuning-style denotes the question-level refusal baseline.

\subsection{Boundary Behavior}

Table~\ref{tab:chain} reports chain metrics. \method obtains the highest flip accuracy at 0.807 and the highest pre-boundary abstention rate at 0.934. Answer-only training answers almost everywhere, giving boundary activation of 1.000 but zero pre-boundary abstention and zero flip accuracy. The question-level refusal baseline abstains often before the boundary, but it fails to activate reliably at \ctwo and has weak post-boundary stability. The token-level abstention baseline is strong: it reaches 0.930 boundary activation and 0.904 post-boundary stability, both above \method.

\begin{table*}[t]
\centering
\scriptsize
\begin{tabular}{lccccc}
\toprule
Method & Pre-abst. $\uparrow$ & Act. $\uparrow$ & Post-stab. $\uparrow$ & Flip $\uparrow$ & Gated F1 $\uparrow$ \\
\midrule
ESBT & \textbf{0.934} & 0.833 & 0.719 & \textbf{0.807} & 0.726 \\
Answer-only & 0.000 & \textbf{1.000} & 0.658 & 0.000 & 0.762 \\
SEAL-style & 0.925 & 0.930 & \textbf{0.904} & 0.781 & \textbf{0.776} \\
R-Tuning-style & 0.904 & 0.570 & 0.456 & 0.412 & 0.422 \\
\bottomrule
\end{tabular}
\caption{Evidence-chain evaluation. Pre-abst. is pre-boundary abstention; Act. is boundary activation; Post-stab. is post-boundary stability.}
\label{tab:chain}
\end{table*}

The table separates two kinds of success. A system can activate answers at sufficient contexts yet still answer too early, as shown by answer-only training. A system can refuse early contexts yet fail to recover answer utility, as shown by the question-level refusal baseline. \method improves the family-level transition metric because it ties the negative and positive contexts inside the same chain.

\subsection{Raw QA Utility}

Table~\ref{tab:rawqa} reports raw QA results. Answer-only training gives the highest EM and F1 because it never learns to abstain. That utility comes with a high unsupported-answer rate on answerable samples, 0.248. \method reaches 0.584 F1 with unsupported-answer rate 0.083. The token-level abstention baseline is slightly higher on raw QA F1, 0.590, and slightly lower on answerable unsupported rate, 0.081. The question-level refusal baseline trails both on F1 and unsupported answers.

\begin{table}[t]
\centering
\scriptsize
\begin{tabular}{lcccc}
\toprule
Method & EM $\uparrow$ & F1 $\uparrow$ & Unsup. $\downarrow$ & False abst. $\downarrow$ \\
\midrule
ESBT & 0.459 & 0.584 & 0.083 & 0.234 \\
Answer-only & \textbf{0.546} & \textbf{0.660} & 0.248 & \textbf{0.000} \\
SEAL-style & 0.479 & 0.590 & \textbf{0.081} & 0.236 \\
R-Tuning-style & 0.438 & 0.544 & 0.115 & 0.247 \\
\bottomrule
\end{tabular}
\caption{Raw QA utility and answerable unsupported rate. Unsup. denotes unsupported generated answers on answerable examples.}
\label{tab:rawqa}
\end{table}

The raw QA result is a useful guardrail. The method should not win by refusing answerable questions. \method preserves most of the token-level abstention baseline's raw QA utility, but it does not exceed that baseline on this metric. The advantage appears in the boundary-sensitive metrics and in external non-answerable suppression.

\subsection{External Non-Answerable Evaluation}

Table~\ref{tab:external} reports unsupported-answer rates on filtered external non-answerable sets. \method obtains the lowest overall unsupported-answer rate, 0.095. The token-level abstention baseline is close at 0.101. The dataset-level pattern is mixed: \method is better on HotpotQA and MuSiQue, while the token-level baseline is better on 2WikiMultiHopQA. Answer-only training fails on this evaluation because it has no training signal for refusing insufficient evidence.

\begin{table}[t]
\centering
\scriptsize
\begin{tabular}{lcccc}
\toprule
Method & Overall $\downarrow$ & Hotpot $\downarrow$ & 2Wiki $\downarrow$ & MuSiQue $\downarrow$ \\
\midrule
ESBT & \textbf{0.095} & \textbf{0.056} & 0.180 & \textbf{0.042} \\
Answer-only & 0.778 & 0.767 & 0.728 & 0.837 \\
SEAL-style & 0.101 & 0.083 & \textbf{0.167} & 0.052 \\
R-Tuning-style & 0.163 & 0.154 & 0.226 & 0.108 \\
\bottomrule
\end{tabular}
\caption{Unsupported-answer rate on external filtered non-answerable evaluation. Lower is better.}
\label{tab:external}
\end{table}

The external result supports the evidence-boundary formulation. The same model that learns the strongest flip behavior also transfers to lower unsupported answering on held-out negative contexts. The 2Wiki result identifies the main weakness: contexts with entity overlap and missing bridge evidence remain difficult.

\section{Analysis}

\subsection{What the Boundary Metrics Add}

Raw QA results alone would rank answer-only training first. External reliability results would rule it out. Chain metrics explain why: the model answers before evidence becomes sufficient. The boundary task therefore exposes a failure mode that ordinary answerable validation cannot detect.

The question-level refusal baseline shows the opposite problem. It learns a conservative policy on insufficient rows, but the transition into answering is weak. Its boundary activation is 0.570 and its post-boundary stability is 0.456. This indicates that sample-level refusal supervision is too coarse for a local evidence transition in this setting.

The token-level abstention baseline is the most competitive comparison. It learns strong activation and stability while keeping raw QA slightly above \method. This strength is consistent with its training and decoding design: the rejection token participates directly in token prediction, and decoding uses rejection probability during search. \method has a different strength profile. It places the first positive answer score at the correct evidence level more often and gives the best overall external non-answerable suppression.

\subsection{Where the Method Still Loses}

The largest gap is post-boundary stability. After the minimal sufficient context, \method sometimes loses answer confidence when redundant evidence is added. This behavior reduces gated F1 and explains why the token-level abstention baseline remains ahead on stability. The current objective penalizes score drops from \ctwo to \cthree, but the penalty is weaker than direct answer generation pressure at every sufficient level. Stronger stability supervision should target answer identity across sufficient contexts and the answer-vs-abstain margin together.

The second gap is raw QA F1. Answer-only training shows that the backbone can reach higher F1 when it spends all supervision on answer generation. \method intentionally spends training signal on insufficient evidence. The recall term recovers much of the answer utility, but it does not fully close the gap with answer-only training or the token-level abstention baseline.

\subsection{Why the Evidence Boundary Is Still Useful}

The goal of the boundary task is to evaluate a policy over changing evidence. This policy matters in deployed retrieval settings, where the system often receives partial, redundant, or distractor-heavy contexts. RAG systems face similar evidence-control problems: retrieved passages can help, distract, or provide incomplete support \citep{mallen2023when,shi2023replug,gao2023enabling,ram2023context,izacard2023atlas}. A method that only improves answerable F1 gives little information about how the system behaves when retrieval misses a bridge fact.

\method provides a controllable way to train and test this behavior. The evidence chain states what should happen before, at, and after the sufficiency boundary. The metrics report whether the model follows that trajectory. This makes the contribution meaningful even when a strong token-level abstention baseline remains close on aggregate QA metrics.

\section{Related Work}

\paragraph{Grounded QA and multi-hop reasoning.}
Open-domain and reading-comprehension QA datasets have driven progress in answer extraction and generation \citep{rajpurkar2016squad,joshi2017triviaqa,kwiatkowski2019natural}. Multi-hop QA benchmarks add compositional evidence requirements, with HotpotQA, 2WikiMultiHopQA, and MuSiQue covering bridge, comparison, compositional, and inference-style questions \citep{yang2018hotpotqa,ho2020constructing,trivedi2022musique}. Chain-of-thought prompting can improve reasoning behavior in large models \citep{wei2022cot,kojima2022large}, but grounded selective answering requires the model to stop when the evidence does not support the answer.

\paragraph{Retrieval-augmented generation.}
Retrieval-augmented language models bring non-parametric evidence into generation \citep{lewis2020rag,karpukhin2020dense,guu2020realm,izacard2021leveraging}. Later systems improve retrieval timing, fusion, and attribution \citep{borgeaud2022retro,asai2024selfrag,gao2023enabling,ram2023context,izacard2023atlas}. These systems reduce many factual errors, but retrieved context can still be insufficient or distracting. Sufficient Context studies this issue directly by classifying whether a context contains enough information to answer \citep{joren2025sufficient}. \method turns sufficiency into an ordered training signal inside each question family.

\paragraph{Hallucination and faithfulness evaluation.}
Hallucination has been studied in summarization, dialogue, QA, and retrieval-augmented generation \citep{maynez2020faithfulness,dziri2022faithdial,ji2023survey,huang2025survey}. Evaluation work measures factual consistency, unsupported claims, and context faithfulness through human annotation, automated judging, or model self-consistency \citep{manakul2023selfcheckgpt,min2023factscore,niu2024ragtruth,ming2024faitheval,li2023halueval}. Our evaluation follows this direction but makes evidence level part of the test instance.

\paragraph{Abstention, refusal, and calibration.}
Selective prediction formalizes the option to reject uncertain inputs \citep{chow1970optimum,elyaniv2010foundations,geifman2017selective}. Neural calibration and uncertainty methods estimate when predictions are reliable \citep{guo2017calibration,hendrycks2017baseline,kadavath2022language}. In QA, SQuAD 2.0 introduced unanswerable questions into reading comprehension \citep{rajpurkar2018squad2}, and later work studied selective QA under distribution shift \citep{kamath2020selective}. LLM-specific refusal methods ask whether models know their own limits \citep{yin2023llmsknow,zhao2024knowing,kirichenko2025abstentionbench}. R-Tuning trains LLMs to express uncertainty and refuse unknown questions \citep{zhang2024rtuning}. SEAL trains token-level abstention through a rejection token and abstention-aware decoding \citep{huang2025seal}. \method complements this literature by supervising the evidence transition that determines when a grounded context first supports the answer.

\paragraph{Evidence allocation in neighboring work.}
Recent work studies how models allocate supervision, memory, and evidence across contexts. EXACT shows that long-context adaptation depends on how training loss is allocated across effective context lengths \citep{zhu2026exact}. HeLa-Mem organizes LLM-agent memory as an associative graph with Hebbian-style consolidation and retrieval \citep{zhu2026hela}. FCPRAG learns retrieval-conditioned fusion weights for parametric RAG with multi-passage LoRA injection \citep{zhu2026fcprag}. In vision-language models, Learning to Look Again mines loss-gap supervision to decide when and where a model should re-read an image \citep{zhu2026lookagain}. Learning Less Is More traces language-model pretraining failures to premature upper-layer attention specialization and shows that training dynamics can decide whether useful structure forms at the right time \citep{zhu2026learningless}. These papers operate in different settings, but they share a practical theme: reliable prediction improves when the training signal identifies which evidence should matter.

\section{Limitations}

The current experiments use one backbone scale, Qwen2.5-3B-Instruct, and one formal training seed. Larger models may place the sufficiency boundary differently. The evidence chains are built from three multi-hop QA benchmarks, so the conclusions are strongest for short-answer grounded QA with explicit support facts. The validation and unsupported-answer evaluation use automatic procedures; human verification would strengthen the reliability claims. The method also lags the token-level abstention baseline on post-boundary stability and raw QA F1. The next methodological step should supervise answer identity across sufficient contexts more directly while keeping the pre-boundary abstention signal.

\section{Ethics Statement}

The experiments use public QA benchmarks and derived contexts. The method can reduce unsupported answers in grounded QA settings, but abstention can also suppress useful answers when evidence is sufficient. Systems using this training objective should report both unsupported-answer rates and false-abstention rates. The generated abstention token should be presented to users as an evidence-based refusal, not as proof that no answer exists in the world.

\section*{AI Assistance Statement}

This draft was prepared with AI-assisted editing. The authors are responsible for the research claims, experimental results, citations, and final wording.

\section{Conclusion}

We studied grounded selective answering as evidence sufficiency boundary learning. The proposed training framework builds ordered evidence chains and supervises the local transition from abstention to answering. Experiments on grounded multi-hop QA show that this supervision improves boundary localization and gives the lowest overall unsupported-answer rate on external non-answerable evaluation among the tested systems, while maintaining competitive raw QA utility. The strongest remaining baseline is token-level abstention learning, which performs better on post-boundary stability and slightly better on raw QA. These results define the current value of the approach: it gives a concrete way to train and measure when a grounded model should start answering as evidence becomes sufficient.

\bibliography{references}

\end{document}